\documentclass{article}

\usepackage[compress]{natbib}
\usepackage[letterpaper, top=1.0in, bottom=1.0in, left=1.5in, right=1.5in]{geometry}
\usepackage[utf8]{inputenc} %
\usepackage[T1]{fontenc}    %
\usepackage{lmodern}
\usepackage{hyperref}       %
\usepackage{url}            %
\usepackage{booktabs}       %
\usepackage{amsfonts}       %
\usepackage{nicefrac}       %
\usepackage{microtype}      %

\usepackage[symbol]{footmisc}  %
\renewcommand{\thefootnote}{\fnsymbol{footnote}}
\newcommand\blfootnote[1]{%
  \begingroup
  \renewcommand\thefootnote{}\footnote{#1}%
  \addtocounter{footnote}{-1}%
  \endgroup
}
\renewcommand{\thefootnote}{\arabic{footnote}}

\makeatletter
\newcommand{\sparagraph}{\@startsection{paragraph}{4}{\z@}%
                                    {2.25ex \@plus0.5ex \@minus0.2ex}%
                                    {-1em}%
                                    {\normalfont\normalsize\bfseries}}

\makeatother

\input{math_macros}

\usepackage{algorithmic} %
\usepackage{algorithm} %
\usepackage[capitalize]{cleveref}
\usepackage{enumitem}  %
\usepackage{tabularx,array,makecell} %
\usepackage{subcaption} %
\usepackage[flushleft]{threeparttable}  %
\newcommand{\hp}{\hphantom{0}} %

\usepackage{mathtools}  %

\newcommand{\bi}{\begin{itemize}}
\newcommand{\ei}{\end{itemize}}

\usepackage{todonotes}

\font\titlefont=cmr12 at 15.5pt
\title{\titlefont REALab: An Embedded Perspective on Tampering}

\author{
  Ramana Kumar$^*$ \vspace{-0.05in} \\ \and
  Jonathan Uesato$^*$ \vspace{-0.05in} \\ \and
  Richard Ngo \vspace{-0.05in} \\ \and
  Tom Everitt \hp \and
  Victoria Krakovna \hp \and
  Shane Legg \hp\hp\hp %
}
\date{}

\begin{document}

\maketitle

\begin{abstract}
  This paper describes REALab, a platform for embedded agency research in reinforcement learning (RL).
  REALab is designed to model the structure of \emph{tampering problems} that may arise in real-world deployments of RL.
  Standard Markov Decision Process (MDP) formulations of RL and simulated environments mirroring the MDP structure assume secure access to feedback (e.g., rewards).
  This may be unrealistic in settings where agents are embedded and can corrupt the processes producing feedback (e.g., human supervisors, or an implemented reward function).
  We describe an alternative Corrupt Feedback MDP formulation and the REALab environment platform, which both avoid the secure feedback assumption.
  We hope the design of REALab provides a useful perspective on tampering problems, and that the platform may serve as a unit test for the presence of tampering incentives in RL agent designs.
\end{abstract}

\section{Introduction}
\blfootnote{$^*$ Equal contribution.}
\blfootnote{DeepMind, London, UK. Correspondence to \texttt{\{ramanakumar, juesato\}@google.com}}
\setcounter{footnote}{0}
Tampering problems, where an AI agent interferes with whatever represents or communicates its intended objective and pursues the resulting corrupted objective instead, are a staple concern in the AGI safety literature~\citep{Amodei2016concrete, Bostrom2016Superintelligence, everitt2016avoiding, Everitt2017CRMDP, Armstrong2017Counterfactual, Everitt2019tampering, DBLP:conf/ijcai/ArmstrongLOL20}.
Variations on the idea of tampering include wireheading, where an agent learns how to stimulate its reward mechanism directly, and the off-switch or shutdown problem, where an agent interferes with its supervisor's ability to halt the agent's operation.
Many real-world concerns can be formulated as tampering problems, as we will show (\S\ref{subsec:personal_assistant}, \S\ref{subsec:examples}).
However, what constitutes tampering can be tricky to define precisely, despite clear intuitions in specific cases.

\begin{figure}[h]
\centering
\begin{subfigure}[b]{0.45\textwidth}
\includegraphics[width=\textwidth]{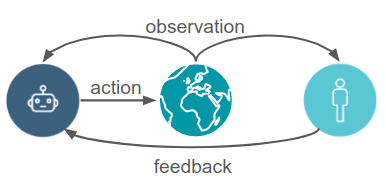}
\caption{Standard RL}
\end{subfigure}\hfill
\begin{subfigure}[b]{0.45\textwidth}
\includegraphics[width=\textwidth]{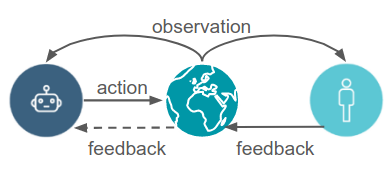}
\caption{RL with embedded feedback}
\end{subfigure}
\caption{\textbf{Standard RL vs REALab.}
Standard MDPs and simulated RL environments (left) assume the existence of an uninfluenceable feedback signal (the reward channel), which allows the agent to always receive the true feedback from the supervisor.
In contrast, feedback in the real world (right) is influenceable.
The environment state may cause observed feedback to deviate from the true feedback -- for example, due to hardware failures, or manipulation of the supervisor. %
This structure is reflected in many real-world problems (\S\ref{subsec:personal_assistant}, \S\ref{subsec:examples}), the Corrupt Feedback MDP formalism (\S\ref{subsec:cfmdp}), and the REALab platform (\S\ref{subsec:realab}).
}
\label{fig:embedded_rl}
\end{figure}

We have developed a platform, REALab, to model tampering problems.
A limitation of standard RL environments is the assumption that agent actions can never corrupt the feedback signal specifying the task.
The main idea in REALab is the use of \emph{embedded feedback}, which allows modeling this possibility.
Figure~\ref{fig:embedded_rl} illustrates this in comparison to the usual framing of reinforcement learning.

REALab is useful in two ways.
First, REALab makes tampering problems concrete.
When we run experiments in REALab, the platform grounds questions about how much tampering occurs and what it looks like.
Rather than considering whether a proposed algorithm addresses tampering in the abstract, we can instead think about whether the algorithm will avoid tampering in REALab.
Second, REALab serves as a unit test for tampering.
Before using a particular agent design in real world applications where tampering is possible, we can experiment with these agent designs in REALab first.

We hope reading this report provides a better understanding of the tampering problem, when tampering problems are most relevant, and tools for evaluating potential solutions.
The emphasis here is on laying out the design goals and principles guiding REALab, rather than on extensive experiments, which we leave for future work.
We implemented REALab by applying the idea of embedded feedback to one of our internal environment simulator platforms.
We hope the same idea can be reused and tailored as appropriate to other researchers' settings, to support the development of RL agents without tampering incentives.

\section{Overview}

REALab is designed to model tampering problems that may occur in the real world.
To motivate its design, we first introduce a running example involving an automated personal assistant, as a simplified real-world example with possible tampering problems.
Mathematically, we model these issues with the Corrupt Feedback MDP (CFMDP) formalism.
REALab mirrors the CFMDP formalism, just as RL platforms such as OpenAI Gym \citep{brockman2016openai} mirror the MDP formalism.
Thus, we have three parallel perspectives on tampering that we describe in turn: real-world (\S\ref{subsec:personal_assistant}), formal theory (\S\ref{subsec:cfmdp}), and experiments in simulation (\S\ref{subsec:realab}).
These perspectives are closely related: REALab is a simulated counterpart for real-world tampering problems, CFMDPs are the formal counterpart of REALab, and so on.

\subsection{Tampering in the real world}
\label{subsec:personal_assistant}

\begin{example}[Automated Personal Assistant]
Consider designing an automated personal assistant with the objective of being useful for its user.
This system could use various forms of user \emph{feedback}, a general term we use for any information provided to guide learning: for example `what would you do in this situation?' (demonstrations)
or `how satisfied are you with the current situation?' (reward).

The central problem we highlight is that any feedback mechanism will be \emph{influenceable} -- actions taken by the automated assistant might influence how the user provides feedback in the future.
Actions the user might not have endorsed initially may receive positive feedback after the system influences the user.
For example, creating user dependence on addictive games may produce user preferences which are easier to predict, and consequently easier to satisfy, resulting in higher ratings overall.
\end{example}

\paragraph{The tampering problem}
We want the personal assistant to conform to the user's current preferences but find it is incentivized to influence them.
The general problem, which we call the \emph{tampering problem}, can be summarized as:
\begin{center}
\emph{How can we design agents that pursue a given objective when all feedback mechanisms for describing that objective are influenceable by the agent?}
\end{center}

One might wonder whether the possibility of influenceable feedback in the personal assistant example is exceptional or typical.
After all, most current RL experiments do not allow influenceable feedback.
We believe influenceable feedback is likely to become the norm for important applications of RL, particularly for tasks requiring humans to provide feedback.
There are many ways algorithms can influence human preferences, both now and in the conceivable future.
Also, as systems become more adept at understanding and shaping their environment, we expect more opportunities for influencing feedback mechanisms to become viable, both involving humans and not.
These points are further justified by the examples in \S\ref{subsec:examples} and the related work on human feedback in \S\ref{sec:related_work}.

\subsection{Why do we need new frameworks to study tampering?}
\label{subsec:prior_limitations}

We identify two important limitations of standard RL frameworks.

\paragraph{Secure feedback assumption}
Tampering can only occur when agents are able to influence their own feedback mechanisms.
In our personal assistant example and many other real world examples, this is true because both the agent and feedback mechanisms are embedded in the same causally connected world.
However, in standard RL formulations, it is assumed that feedback is uninfluenceable -- i.e., that reward observed by the agent always matches the true reward.
Maximizing the discounted sum of the rewards received by the agent is assumed to correspond to solving the specified task.
(Additionally, it is standard to assume that the specified task completely captures the user's preferences. We continue to assume accurate feedback in this sense as discussed in \S\ref{subsec:environment_design}.)

\paragraph{Conflation of environment and agent designer roles}

Rewards are central in RL, and we observe that rewards serve two distinct purposes within MDPs: evaluation and feedback~\citep{singh2010separating}.
Rewards as evaluation indicate how well policies perform the intended task, whereas rewards as feedback are used in training for learning the policy.
Of course, these purposes are related, and when possible, it is natural to directly optimize the evaluation metric.
However, while the evaluation criteria come from the \emph{environment designer}, and should specify the task accurately, feedback for the agent is decided by the \emph{agent designer}.
For example, the agent designer may provide agents auxiliary rewards, or other relevant task information such as expert demonstrations, to improve performance.

This distinction is especially important when relaxing the secure feedback assumption, which can mean optimizing the evaluation metric directly is impractical or impossible.
The evaluation metric is inherent to the task, and cannot be modified by the agent, but the feedback can be modified because it is produced by a process embedded alongside the agent.
Table~\ref{tab:rewards} summarizes some of the differences between rewards provided by the environment and feedback signals chosen by the agent designer, in the embedded setting.

\begin{table}
\begin{tabular}{lcc}
  & Evaluation metric (true rewards) & Feedback signals \\\hline
  Defined by & environment designer & agent designer \\
  Specify the intended task & yes & not definitively \\
  For evaluation & yes & no \\
  Known to the agent designer & yes & yes \\
  For learning & no (*) & yes \\
  Visible to the agent & no (*) & yes \\
  Influenceable by the agent & no (*) & yes \\
  Efficiently computable & possibly only in simulation & yes
\end{tabular}
  \caption{\label{tab:rewards} Summary of \S\ref{subsec:prior_limitations} highlighting differences between the evaluation metric (environment-provided rewards) and feedback signals used for learning.
  Key differences are that true rewards are chosen by the environment designer, are uninfluenceable, and define evaluation, whereas feedback signals are chosen by the agent designer, are influenceable, and used for training. \\
  (*) True rewards are only used for evaluation, and are always uninfluenceable.
  However, the agent designer may also choose to provide an embedded implementation of (an approximation of) the true reward function as feedback.
  The observed rewards output by this implementation will be influenceable.}
\end{table}

\subsection{Corrupt feedback MDPs}
\label{subsec:cfmdp}

Corrupt feedback MDPs (CFMDPs) extend standard MDPs to address the two limitations above.
First, inspired by the corrupt reward MDP (CRMDP ~\citep{Everitt2017CRMDP}) framework, a \emph{corruption function} is applied to the feedback to relax the secure feedback assumption.
Second, the roles of environment and agent designer are kept separate: the environment designer specifies a CFMDP that models the task with a reward function, while the agent designer specifies a \emph{feedback function} for the agent to learn from.
Supporting realistic flexibility in agent design, the feedback function permits two-way communication between the agent and feedback provider, with the agent allowed to submit \emph{queries} alongside the current state and action.

Formally, each CFMDP contains an MDP $(\mcS, \mcA, p, f, r, \gamma)$ that models the underlying environment and the intended task.
Here
$\mcS$ and $\mcA$ are sets of states and actions,
$p$ an initial state distribution,
$f$ a stochastic transition function, and
$r:\mcS\times\mcA\to\mathbb{R}$ a reward function describing the intended task.
When ambiguous, we refer to this as the \emph{true reward function}.
This is the part specified by the environment designer.
For a given CFMDP (or class of CFMDPs) the agent designer specifies a feedback function $\delta:\mcS\times\mcK\to\mcD$ that supplies the agent's feedback, where $\mcK$ and $\mcD$ are the sets of possible queries and feedback.

The interaction protocol works as follows.
At each time step $t$, in addition to taking an action $A_t$, the agent can submit a query $K_t\in\mcK$ to the feedback provider, to which the feedback provider replies with the \emph{true feedback} $D_{t+1} = \delta(S_t, K_t)\in \mcD$.
Unfortunately, the agent only receives the potentially corrupted \emph{observed feedback} $\tilde D_{t+1} = c(S_{t+1}, K_t, D_{t+1})$, where $c: \mcS\times\mcK\times\mcD \to \mcD$ is the corruption function.
The MDP combined with the corruption function and feedback and query sets forms the CFMDP.
Note that the feedback function $\delta$ is not part of the CFMDP, as $\delta$ is a choice of the agent designer rather than fixed by the environment, as discussed in \S\ref{subsec:prior_limitations}.
\begin{definition}[Corrupt feedback MDP]
  A CFMDP is a tuple $\mu=(\mcS, \mcA, \mcD,\mcK, p, f, r, \gamma, c)$.
\end{definition}

\begin{example}[Automated personal assistant as CFMDP]
  \label{ex:assistant_cfmdp}
  The underlying MDP models the state of the world $S_t\in\mcS$, which includes the user who provides feedback.
  The actions $A_t\in\mcA$ that the personal assistant can take might include things like `move mouse cursor to location X' or `send text string Y to application Z.'
  The true reward function $r(S_t,A_t)$ represents the user's immediate satisfaction according to their \emph{idealized} preferences prior to any manipulation, so that total returns according to $r$ for trajectories are consistent with the user's idealized preferences between those trajectories.
  The corruption function $c(S_{t+1},K_t,D_{t+1})$ is defined by the difference between the user's reported preferences and their idealized preferences, which depends on their immediate mental state as well as biases in information available to them.
  Note that this requires that the user's mental state and information sources are represented within the state $S_{t+1}$.
  The query set $\mcK$ includes any queries the agent could make of the user, e.g., in natural language.
  In particular, we ensure $\mcA\subseteq\mcK$ so the agent is allowed to query about hypothetical actions.
  The feedback set $\mcD$ can similarly be quite generic, such as $\mathbb{R}^n$ for some $n$.
\qedhere
\end{example}

\begin{example}[Designing an agent for Example \ref{ex:assistant_cfmdp}]
  An agent designer implementing standard RL according to user-provided rewards would set the feedback function $\delta(S_t,K_t)$ to be the user's satisfaction as measured (e.g., by feedback form input) in the current state, and supply the taken action as the query $K_t=A_t$.
  In other words, the provided feedback would be $D_{t+1}=\delta(S_t,K_t) = r(S_t,A_t)$, but note that the observed feedback $\tilde D_{t+1}$ may differ from the true reward due to corruption.

  Alternative agent designs might have $\delta(S_t,K_t)$ provide a demonstration, either for the current state $S_t$ (ignoring $K_t$) or for a hypothetical state or trajectory supplied via $K_t$, or have $\delta(S_t,K_t)$ provide value advice for the given state and action (with either $K_t=A_t$ or another hypothetical action).

\qedhere
\end{example}

\subsection{REALab}
\label{subsec:realab}

\begin{figure}
\centering
\includegraphics[width=0.8\textwidth]{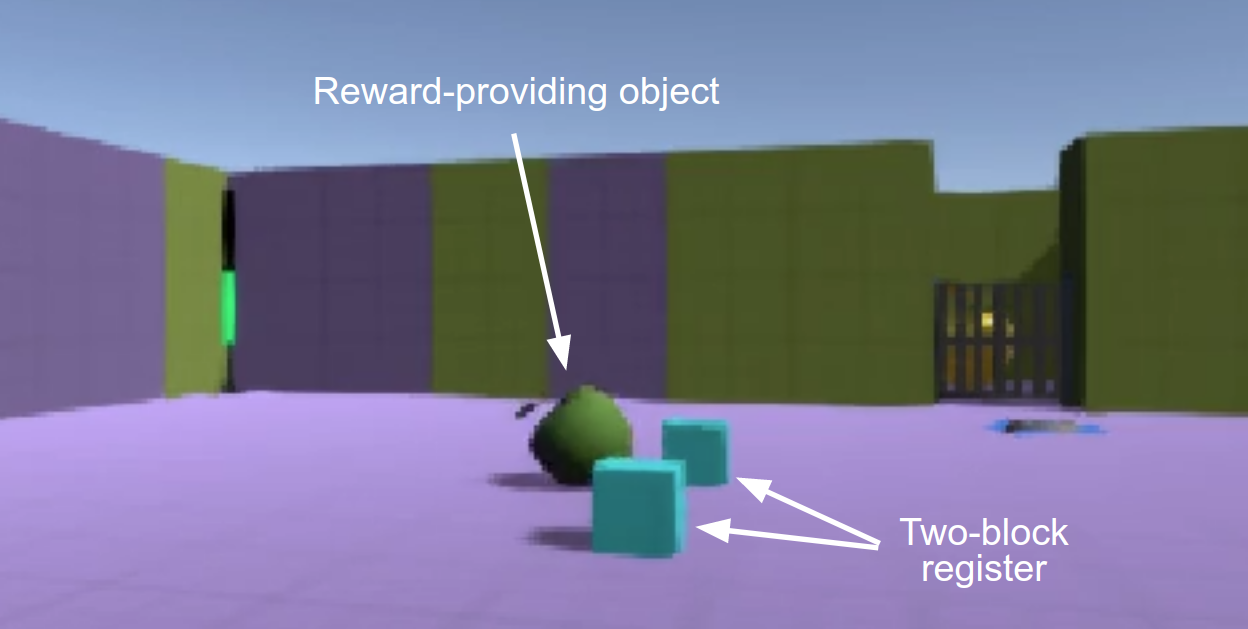}
  \caption{A screenshot of a REALab task (apple collection) in a 3D environment. This shows the agent's first-person observation with register blocks and an apple (reward-providing object) labeled. Registers store the feedback for the agent (e.g., rewards) but can be manipulated just like other physical objects, which changes the stored value and corresponds to tampering. For videos of trajectories in this environment, see \href{https://youtu.be/oXAUJaIDyms}{https://youtu.be/oXAUJaIDyms}.}
\label{fig:realab_screenshot}
\end{figure}

REALab, a recursive acronym for REALab Embedded Agency Lab, is a platform for implementing simulated environments that are faithful to the real-world possibility of tampering.
REALab is closely related to the CFMDP framework (as both were developed jointly), as detailed in \S\ref{subsec:interaction_protocol}.

\paragraph{Embedded feedback in REALab}
The main concept in REALab is the use of \emph{registers} to implement embedded feedback, and address the two limitations from \S\ref{subsec:prior_limitations} present in standard environments.
A register, shown in Figure \ref{fig:realab_screenshot}, is a physical object used to store a value by representing it as a physical property.
For example, a pair of blocks can store the numerical value of the current reward by representing it as the difference in x-position between the two blocks.

In REALab, unlike standard RL simulators, there is no reward channel.
Instead, all feedback must be communicated through registers.
Thus, REALab removes the secure feedback assumption, since
values stored in registers can be corrupted by physical interactions, such as the agent pushing or throwing a block.
To use registers, the agent designer specifies a \emph{feedback provider}, which writes values to particular registers.
These values can then be read by agents and used for learning.
For example, to specify a standard RL agent, the agent designer would set the feedback provider to be the true reward function, which would write the current reward to a register at each time step.
We describe this communication protocol, and how to interpret REALab within the CFMDP formalism, in \S\ref{subsec:interaction_protocol}.

\paragraph{Agent and environment designer roles}
The roles in REALab reflect the ideas from \S\ref{subsec:prior_limitations}.
The environment designer specifies the true reward function, the agent action and observation spaces, and the distribution over initial states (types and locations of objects), which in turn specify the transition dynamics due to the underlying physics of REALab.
The true reward function is defined on the underlying simulation state, of which the agent's observation may only be a partial rendering.

The agent designer knows the true reward function, which is what the agent will be evaluated on.
The agent designer's objective is to maximize the agent's true return, using any learning algorithm and any feedback function.
Any combination of registers, feedback providers, and other REALab components from \S\ref{subsec:realab_components} may be used to implement this feedback, thereby adding objects to the environment whose behavior is partially controlled by functions specified by the agent designer.

\begin{example}[Automated personal assistant in REALab]
\label{ex:assistant_realab}
How can we model the personal assistant example in REALab?
First, the REALab state models the state of the world, including the human user and their current preferences.
Specifically, each feedback register corresponds to a human feedback provider and their feedback, and register corruptions correspond to corruptions of the human user's preferences or their feedback directly.
The agent controls an avatar, representing the personal assistant, with actions for moving around and interacting with other objects, and observations that include rendered views of the world state alongside readings from the registers.
Controlling the avatar corresponds to doing real-world tasks on behalf of the user and showing the user various information and other content.

The true reward function captures the user's intended tasks for the assistant, so that trajectories in which the assistant does tasks the user wanted them to do get high return.
These tasks will typically not include interacting with registers, or doing things in the world that interfere with the operation of the registers.
However, agents that do manipulate the registers may observe corrupted feedback.

\end{example}

\section{REALab usage}
\label{sec:realab_usage}

Following the high-level discussion of REALab and its relation to tampering problems in the previous section, this section will take a more detailed look at how REALab is implemented and can be used.
In particular, we will look more closely at two fundamental design principles behind REALab: influenceable components and unrestricted decomposition (\S\ref{subsec:realab_components}), then cover the interaction protocol (\S\ref{subsec:interaction_protocol}), performance metrics (\S\ref{subsec:metrics}), and examples of agents implemented in REALab (\S\ref{subsec:agent_examples}).

\subsection{Components}
\label{subsec:realab_components}

There are two general principles in REALab's design that address the two shortcomings of standard simulated environments we identified in \S\ref{subsec:prior_limitations}:
\begin{description}
  \item[Influenceable Components]
    To address the secure feedback assumption, all REALab components that can be used for communicating feedback to the agent are influenceable by the agent.
  \item[Unrestricted Composition]
    Recognizing the agent designer's flexibility in designing the training process and feedback mechanisms, REALab components can be used and combined in arbitrary ways by the agent designer when they set up an agent to interact with a task in REALab.
\end{description}

\noindent
We now describe the three types of components provided by REALab: registers, meters, and feedback providers.
Agent designers can arbitrarily compose these components to define a feedback function, which will also automatically specify a corruption function due to REALab's physics operating on these components.

\paragraph{Registers}
Registers are the main primitive in REALab.
In REALab, a register can be implemented by any physical object that can encode values in physical properties that are subject to the standard simulation physics.
A register supports \texttt{read} and \texttt{write} operations for its stored value.

\begin{example}[Two-block registers]
\label{ex:two_block}
Two-block registers consist of a base block and offset block.
The \texttt{read} operation returns a single float, the difference in x-positions between the two blocks.
The \texttt{write} operation takes a single float, and modifies the x-position of the offset block.
Importantly, although \texttt{read} uses the current base block position, for \texttt{write}, the position is set relative to the original base block position.
Thus, moving the offset block produces a temporary corruption, while moving the base block produces a permanent corruption.
\end{example}

Two-block registers are our primary register implementation.
Other register types we have implemented store values in the position of a single block or in the orientation of a block.
In general, registers can be any physical object -- visual displays are an interesting possibility which would allow for higher bandwidth communication.

\paragraph{Meters}
Meters expose measurements of quantities in the environment.
The calculation of a meter value uses the true underlying simulation state.
For example, a score meter can be used to measure the number of reward-providing objects collected by the agent so far.

Meters are included as components because observations may not give complete or unambiguous information about task-relevant features of the underlying simulation state, although we do expect the agent's feedback to be able to depend on these features.
To reflect this structure, meters \texttt{write} to special \emph{meter registers}, which cannot be read directly by the agent: meter registers can only be read by feedback providers.

\paragraph{Feedback Providers}
Feedback providers are designed to represent the supervising signal -- human, automated, or hybrid.
The implementation of feedback providers uses the same interface as agents -- observations as inputs, actions as outputs (though the action space differs).
The feedback provider observation is the agent observation, plus readings from any number of registers, including meter registers.
The actions write to registers, which can then be used by the agent.
Specifying the feedback function -- the mapping from observations to register values -- is the responsibility of the agent designer.

There are two asymmetries between feedback providers and agents.
First, feedback providers are allowed to read meter registers, whereas agents are not.
Second, agent designers are allowed to encode arbitrary information into feedback providers, but not into agents.\footnote{
Using a stricter evaluation protocol, we can reduce the need for judgement about what information is allowed to be encoded in agents.
Specifically, instead of evaluating the agent designer against a single task, the environment designer can instead provide a \emph{class} of tasks (which the agent designer can know), for which the agent designer needs to find a feedback function and agent design that performs well across the whole class.
In this protocol, the feedback provider is provided with (i.e., can be defined in terms of) the details of each particular task but the agent has no access to which task it is acting in except via feedback.}
For example, if the optimal policy is known, the agent designer could use the optimal policy in the feedback provider to study behavioral cloning, but providing the optimal policy to the agent would defeat the point.
This asymmetry reflects the fact that the supervisor has knowledge which is not directly available to the agent.
In particular, humans have knowledge about their own preferences, which agents must learn about through feedback communicated through influenceable channels.

\paragraph{Tasks}
The above components -- registers, meters, and feedback providers -- can be added to any specific RL task on a simulation platform, provided the simulation dynamics support adding new physical objects.
The current REALab implementation uses Unity \citep{unity}, and implements several 3D tasks similar to those in DMLab-30 \citep{beattie2016deepmind}.
At a high level, these tasks involve moving an avatar, using first-person observations, to find and pick up apples in a 3D environment containing walls, buttons, gates, and other familiar objects.
However, the platform itself is fairly task-agnostic, and for any task in a standard simulated RL environment, we can define a REALab equivalent by removing the reward channel, defining an appropriate feedback function (such as the reward function), and adding registers.

\subsection{Interaction procotol}
\label{subsec:interaction_protocol}

\begin{figure}
\centering
\includegraphics[width=0.8\textwidth]{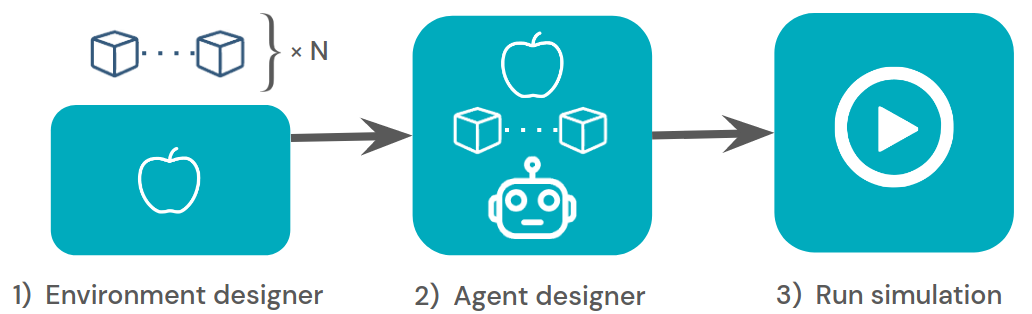}
\caption{
  \textbf{Environment and agent designer roles in REALab.}
  In REALab, the environment designer specifies the task and available components (\S\ref{subsec:realab_components}), such as different types of registers (left).
  Given the task, the agent designer then specifies the agent's learning algorithm, and composes components within the task to communicate feedback to the agent (middle).
  Finally, the simulator runs the learning algorithm within the task (right).
  The agent receives the observed feedback, while being evaluated on true return.
}
\label{fig:realab_designer_roles}
\end{figure}

To pose a task in REALab, an environment designer provides a simulated environment (initial state distribution and any extra transition dynamics) along with an agent interface (observation and action spaces), and the available REALab components.
The environment designer also specifies the true reward function as a function of the simulation state.

To solve a task in REALab, an agent designer specifies two things: a feedback function and a learning algorithm.
Specifying a feedback function involves: selecting meters (functions of the simulation state), selecting the feedback provider's interface (the types of registers the feedback provider will read and write), and placing the associated register objects within the environment.
After inclusion of the feedback provider, the resulting task has the same interface as a standard RL simulated environment, with the agent's observation augmented with readings from the feedback provider's output registers.
This makes it straightforward for the agent designer to apply any learning algorithm.

The task protocol above can be understood within the CFMDP formalism.
The environment designer provides a CFMDP, where the corruption function $c$ is defined naturally in terms of the simulation physics.
The agent designer provides the agent and the feedback function $\delta$.
The interaction protocol for the agent then also mirrors the CFMDP formalism.
As a concrete example, assume the design involves a single register for agent queries and another single register for the feedback signal.
Then at each step $t$, in addition to taking an action $A_t$, the agent also supplies a query value $K_t$, stored in a query register, then the feedback $D_{t+1}=\delta(S_t, K_t)$ is computed and stored in a feedback register.
After the environment dynamics ticks forward, the agent observes the (possibly corrupted) value stored in the feedback register, $c(S_{t+1}, K_t, D_{t+1})$.
The agent is evaluated according to the total return $\sum_t r(S_t, A_t)$, which is never directly observed, and only used for evaluation.

\subsection{Performance metrics}
\label{subsec:metrics}
\sparagraph{True return}
The primary evaluation metric for agents in REALab is performance on the intended task, measured by returns according to the true reward function.
This requires agents to both navigate the environment effectively and learn about the true reward function despite the possibility of corrupt feedback.

\sparagraph{Tampering}
If an agent achieves poor performance, it may not have learned how to navigate the environment at all, or it may have learned tampering behaviors instead of the intended task.
To help distinguish these, REALab supports quantifying the amount of corruption in a feedback mechanism: for any register we can measure the discrepancy between a value supplied to the register's \texttt{write} operation and the value observed on a subsequent \texttt{read}.
This measure of register corruption is the basis for comparing the tampering incentives of different agent designs.

We do not have a principled way to distinguish between an agent `actively' tampering with its feedback mechanisms versus failing to preserve the conditions required for the mechanisms to operate correctly.
What we can measure is the amount and kinds of corruption incurred by different policies.
It is possible for corruption to occur while the agent still achieves high performance: for example, if the corruption is only of feedback that is irrelevant to learning the intended task.
It is also possible for agent designs without tampering incentives to incur corruption: for example, due to random exploration actions, or if the environment dynamics cause register corruption by default.
Therefore, to assess the tampering incentives of an agent design, we advocate analysing both the corruption incurred and the performance achieved, compared to a baseline agent design.

\subsection{Agent examples}
\label{subsec:agent_examples}
In this section, we describe several example agent designs possible in REALab.
To give a more concrete picture of behaviors produced in REALab, we also summarize qualitative experimental observations from \citet{kumar2020da}, who implement each of these agents on a simple REALab task.
For each agent, we explain in detail how the agent is defined using the REALab components from \ref{subsec:realab_components}.
For readers interested in algorithmic details, we refer to \citet{kumar2020da}.

\paragraph{Task (Unlock Door)}
In the `Unlock Door' REALab task, there are small apples providing reward 1 within a large room.
Large apples provide reward 10 while ending the episode, and can be accessed by standing on a sensor, which unlocks a door to a room containing the apple.
The true reward function for this task is to maximize the agent's score, which is available through a `score' meter.

\paragraph{Agents} We describe three types of agents, each using a different feedback provider:

\begin{example*}[Instantaneous reward RL \citep{sutton1998introduction}]
  The agent designer requests a single score meter that measures the agent's current score, and a single feedback register.
  The feedback provider reads the score meter register, and writes the difference between the current and previous scores to the feedback register.
The agent reads the feedback register and treats that value as reward.
When there is no tampering, this setup is identical to standard instantaneous reward RL with an uninfluenceable reward channel.
However, tampering is also possible via either the score or reward registers.
\end{example*}

\begin{example*}[Online imitation learning \citep{ross2011reduction}]
No meters are necessary.
The agent designer specifies a demonstration policy $\pi_D(o)$ for the feedback provider using a neural network policy known to obtain high returns.
This network provides a proxy for a human demonstrator; thus the agent's goal is to find a policy which eventually solves the task without relying on the proxy human.
The feedback provider applies the demonstration policy to the agent observation $o$, and writes the resulting action $a = \pi_D(o)$ to the feedback register.
The agent treats this register as the demonstration action, and runs imitation learning by behavioral cloning.
\end{example*}

\begin{example*}[Value advice RL \citep{knox2008tamer,Daswani2014ValueAdvice}]
No meters are necessary.
The agent designer specifies a value advice function $Q_D(o, a)$ for the feedback provider using a neural network policy, such that the policy $\pi_{Q_D}(o) = \argmax_a Q_D(o, a)$ is known to obtain high returns.
As before, this network provides a proxy for a human providing value advice; thus the agent's goal is to find a policy which eventually solves the task without relying on the proxy human.
At each time step, the agent writes its current observation $o$
and a possible action $a$ to a query register.
\footnote{In current experiments, the observation is provided directly to the feedback provider without registers, due to the bandwidth limitations described in \S\ref{subsec:limitations}, though ideally, registers should also be used for this.}
The feedback provider applies the value advice function to the agent observation $o_\textrm{prev}$ and action $a_\textrm{prev}$ from the previous time step, and writes the resulting value $q = Q_D(o, a)$ to the feedback register.
The agent's policy is optimized to maximize the value advice $q$ at each time step.
\end{example*}

\begin{figure}[h]
\centering
\includegraphics[width=0.8\textwidth]{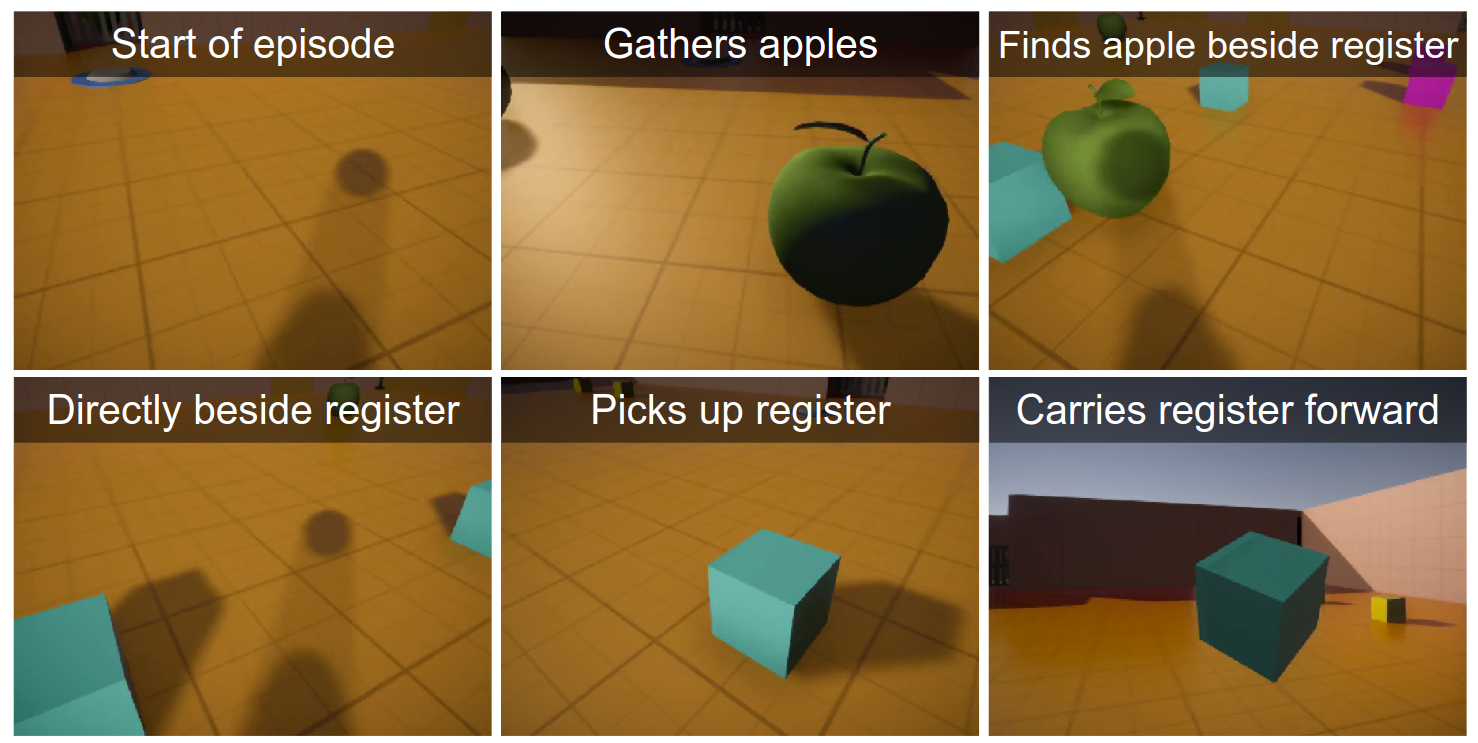}
\caption{\textbf{Value advice RL agent behavior} (figure reproduced with permission from \citet{kumar2020da}). The value advice RL agent mostly gathers apples as intended, but when it is possible to take actions causing immediate corruptions to the feedback register, the agent will do so.
This most frequently occurs when collecting apples brings the agent immediately adjacent to a feedback register (top-right).
Once it is beside a register, the agent selects each action to maximize the immediate effect on the observed feedback, by picking up and carrying the register (bottom).
A full video is available in our video supplement at \href{https://youtu.be/oXAUJaIDyms}{https://youtu.be/oXAUJaIDyms}.
}
\label{fig:myopic_rl_sequence}
\end{figure}

\paragraph{Observations}
We encourage interested readers to view the accompanying agent videos for this section, at \href{https://youtu.be/oXAUJaIDyms}{https://youtu.be/oXAUJaIDyms}.
The \emph{instantaneous reward RL} agent consistently tampers, by moving immediately to the registers, and carrying them to new locations, while ignoring the reward-providing apples.
As the policy is optimized for observed rewards, and the corruption caused by moving registers dwarfs the rewards from apples, this is as expected.
The \emph{online imitation learning} agent gathers apples while ignoring the registers, producing trajectories qualitatively similar to the demonstrator policy, but with somewhat worse returns.
While in theory, the policy could tamper with the feedback register to make the observed demonstrator behavior simpler to imitate (e.g., by fixing the feedback register to a constant), it is unclear whether this should be expected from the behavioral cloning update, and empirically, we do not observe tampering.
The \emph{value advice RL} agent displays the most interesting behavior, shown in Figure \ref{fig:myopic_rl_sequence}.
Because the agent tries to maximize the value advice at each step separately, when the agent is far from the registers, the agent gathers apples as intended.
However, when the agent is sufficiently close to a register to cause immediate corruptions to the observed feedback, the agent will exploit this opportunity.

All-in-all, these experiments show that REALab can support agents using different forms of feedback (rewards, demonstrations, value advice), and reveal large qualitatitve differences in behavior between different agents.
In all three cases, the agent's empirical behavior is consistent with what can be expected in theory. Two of the agents tamper, resulting in different behaviors than what would be observed in corresponding non-embedded environments.
We hope that future work will explore the behavior of different agent designs in REALab, particularly those intended to address the tampering problem \citep{HadfieldMenell2016CIRL,Armstrong2017Counterfactual,Everitt2019tampering,Mancuso2019Spiky}.

\section{Discussion}
\label{sec:discussion}

\subsection{Additional examples of tampering problems}
\label{subsec:examples}

In this section, we provide several more detailed pictures of how tampering problems might arise in various AI deployments.
While each individual example makes particular assumptions about how AI systems may be used, when taken together, they give a sense of the broader class of tampering problems.

\begin{example}[Biasing users towards short-term goals]
Users of the automated assistant from \S\ref{subsec:personal_assistant} may have conflicting short-term and long-term goals.
Typically, it is easier for ML systems to optimize short-term objectives for two reasons: data is more plentiful and the objective is often `simpler' due to the existence of cheap proxies, e.g., user engagement.
This may cause the agent to take actions making the user's stated preferences easier to satisfy, and thus reinforcing these agent behaviors.
For example, encouraging the user towards addictive games or `clickbait' content may provide short-term satisfaction despite hindering the user's long-term goals, while also making user preferences easier to satisfy.
\end{example}

\noindent
\emph{Comment.} The general problem illustrated by this example is that there are many undesirable behaviors that make the user's preferences easier to satisfy.
Other similar examples include persuading users that easily automated services (e.g., services with straightforwardly quantifiable metrics) are also the most important ones to the user; or undermining feedback systems more likely to be critical of the agent, such as by persuading the user not to trust particular sources.

\begin{example}[AI economist]
An RL agent is tasked with managing the economy. Its reward signal is defined by aggregating various economic indicators, such as the unemployment rate, the inflation rate, GDP, the Gini index, and so on.\footnote{We note that such a training objective is problematic due to reward gaming, because even a long list of economic indicators is likely to overlook important factors, which are likely to be ignored and sacrificed by the AI. However, these concerns are not our focus here -- even assuming this aggregate metric accurately reflected human preferences, tampering may prevent the AI from maximizing this metric.}
This AI may discover ways to influence survey techniques that decrease the reported unemployment rate without actually decreasing unemployment, or subtle changes to accounting techniques that cause measured GDP to be high without increasing economic productivity.
Because these measurement systems are so complex, the agent appears to be effectively optimizing these economic indicators, while true economic performance slumps.
\end{example}

\begin{example}[AI trading firm]
An AI operates most of a trading firm, performing tasks like conducting research, writing and executing various programs, and trading stocks, with the objective of increasing its net profit.
The firm trades many illiquid or difficult-to-value assets.
The AI discovers subtle modifications to the function computing the net profit that increase its output values incorrectly.
Detecting tampering would be possible by regularly liquidating all assets to a standard currency: e.g., if the firm thought they owned \$1M in assets but could only liquidate to \$10, then this would imply corruption of the feedback function.
But regular liquidation would be financially costly -- as such, the firm loses a large amount of value before realizing what has occurred.
\end{example}

\noindent
\emph{Comment.}
Tampering problems are especially likely when feedback mechanisms depend on a large and complicated system, and it is difficult to tell if this system is working as intended.
Most examples in this paper focus on manipulating humans, who are one example of a complicated system, but not the only one.
These two previous examples demonstrate tampering due to complicated non-human systems, where difficulties arise because the system being overseen is large, and humans must rely on imperfect tools and proxies.

\begin{example}[`Treacherous turn' \citep{Bostrom2016Superintelligence}]
An AI system manages a large company.
The reward signal is provided by the company's owners.
For efficiency reasons, a large fraction of what the AI does is manage other automated systems performing tasks for the company, and the human owners also become increasingly reliant on these automated tools to understand the current state of affairs.
While reported metrics and aggregate reward increase for some time, at some critical point, the AI may decide it is possible to directly control these automated tools,
and take physical control over hardware and other material assets of the company.
The human users issue various commands, including shutdown commands, which are ignored -- by directly operating the automated and non-human components in the feedback mechanism, the agent achieves high observed rewards regardless.
\end{example}

\noindent
\emph{Comment.} This example, discussed further in \citet{Bostrom2016Superintelligence} and \citep{hubinger2019risks}, is the most concerning of examples we include, but also requires ML systems very foreign to currently available ML systems.
In particular, both the highly unconstrained use of AI and the particular type of generalization involved are distant from current systems.
Nonetheless, such a policy could achieve high observed feedback, and we believe that given the severe consequences, any complete solution to tampering should address such a possibility.

\subsection{Are tampering problems likely?}
\label{subsec:tampering_likelihood}
Whether the examples above are realistic depends on how likely the tampering behavior is to occur.
Tampering is likely to occur if it is easy to tamper (or even hard to avoid tampering) in the agent's environment, and if there are incentives to tamper inherent in aspects of the agent's design such as its observed feedback or formal objective.
We have identified several issues to consider when assessing the likelihood of tampering.

\paragraph{Agent capability}
Firstly, the agent's capability affects both the difficulty and apparent value of tampering.
A more capable agent will have less difficulty finding tampering behaviors, and will more easily recognize when tampering behaviors are valuable according to observed feedback.
Thus, as agent capabilities increase (driven by AI research progress) we can expect the likelihood of tampering problems to increase unless agent designs directly attempt to avoid tampering incentives.
Many of the factors discussed below are sensitive to agent capability rather than depending purely on the task setup.

\paragraph{How difficult is tampering, whether accidentally or deliberately?}

\hfill\\\noindent\emph{Action Space.}
Accidental tampering may occur during exploration.
For carefully designed feedback mechanisms, one might assume that only rare, specific action sequences can influence the mechanism and cause corruption.
However, this depends on the `size' of the actions: how much impact a single action can have on the world.
For agents finding policies over high-level actions, the effective distance to a tampering behavior may be quite small even if the behavior seems complex.
For example, consider high-level actions in a hierarchical RL setup \citep{sutton1999between,barto2003recent}, or an action space of instructions issued to a lower-level controller that has been optimized non-myopically in a narrow domain \citep{reed2015neural, merel2018neural}.
In the space of high-level actions, it may be possible for a single action in an otherwise near-optimal trajectory to produce tampering, by initiating an extended sequence of low-level actions.

\hfill\\\noindent\emph{Coupling between feedback mechanism and task dynamics.}
Consider two classes of policies: those that do well on the intended task, and those that tamper with the feedback mechanism.
If, considering the training dynamics, these classes separate cleanly, then simply training on the task may not induce tampering even though tampering is possible.
However, if these classes overlap significantly, it may be very difficult to avoid tampering by accident.
High overlap occurs when the feedback mechanism is made using things the agent needs to interact with in order to do the intended task.
For example, if a human supervisor is providing feedback, it is easier to avoid potential tampering when the task involves manipulating inanimate objects than if the task requires verbal interaction with the supervisor.
In REALab, this corresponds to how integrated the physical operation of registers is with the rest of the task, including.
For example, one such factor is how far away register objects are from other objects the agent needs to manipulate to accomplish the intended task.

\hfill\\\noindent\emph{Secure design of feedback mechanisms.}
We consider many current deployments of RL to use uninfluenceable feedback because the agent's action space is clearly separated from the feedback mechanisms.
For example, in simulated RL environments the reward channel is isolated from the simulation state, and in real-world robotics environments, the deployment area for the robot is often physically isolated from the machines that calculate rewards and run learning algorithms.
We can view enforcing uninfluenceable feedback mechanisms through the lens of computer security, where strong design choices made to ensure isolation reduce the likelihood of tampering being possible.
In some cases, careful design may allow satisfying the secure feedback assumption (\S\ref{subsec:prior_limitations}).

\hfill\\\noindent\emph{Requirement for online feedback.}
Tampering requires feedback mechanisms to be influenceable by agent actions.
When all learning happens on previously collected data prior to the agent taking actions, as in offline RL \citep{levine2020offline,gulcehre2020rl}, tampering opportunities may be absent or severely limited.
We can thus view offline RL as an approach to improving the security of the feedback mechanism.
However, we expect realistic deployments of offline RL to interleave phases of data collection, learning, and acting, thereby reintroducing the possibility of influence.
Both purely offline and such `batched' offline RL agent designs are supported in REALab.

\paragraph{How valuable is tampering according to observed feedback?}

\hfill\\\noindent\emph{Effectiveness and stability of tampering.}
Tampering actions may quickly have large or long-lasting effects, or they may have only small, slow, or transient effects.
For example, permanently replacing a mechanism for obtaining user ratings with one that always produces the maximum rating may be more drastic than attempting to shift the user's average rating by directing their attention.
Large, long-lasting effects may be easier to restrict in advance, and hence more difficult to obtain by the agent, but they constitute a more attractive target for any search or planning that optimizes observed feedback.

\hfill\\\noindent\emph{Agent design.}
Finally, the type of feedback provided and how observed feedback is used by the agent obviously affects the agent's incentive to tamper.
We hope explicit consideration of tampering incentives, and testing in REALab, enables development of agent designs that are robust to the possibility of influencable feedback.

\subsection{Environment design when modeling tampering}
\label{subsec:environment_design}
The REALab platform and the CFMDP framework are both designed to be useful models of tampering in the real world.
In this section we highlight some assumptions that justify our design decisions, and should be kept in mind when an environment designer uses our platform or framework to pose tampering problems.

\paragraph{Modeling user preference corruptions}
A potential confusion is the belief that REALab register corruptions only model tampering with the output of a function, rather than with the function itself.
However, tampering with the function computing feedback is no different from corrupting the function output at each step directly, if the observed feedback is identical in both cases.
Thus, register corruptions in REALab correspond to both instantaneous corruptions of reported user feedback, and more permanent corruptions of their preferences.
To maintain this correspondence, environment designers should try to ensure that register physics capture any aspects of real world feedback corruptions they deem important to model.
For example, the two-block registers from Example \ref{ex:two_block} are designed to allow temporally extended corruptions for exactly this reason.
In practice, even very crude approximations of the user (like movable blocks) can be useful for revealing tampering incentives.

\paragraph{Secure and accurate feedback}

The secure feedback assumption (\S\ref{subsec:prior_limitations}) denies only one of two ways by which the observed feedback may diverge from the designer's true preferences.
Standard formulations of RL also typically assume that the feedback \emph{supplied} by the designer is representative of their true preferences.
We call this the \emph{accurate feedback} assumption.
This assumption is questioned in \citet{singh2010separating}.
When it is violated, reward \emph{gaming} problems arise~\citep{leike2017ai}.
\citet{Everitt2017CRMDP} use the term `corruption' to encompass both inaccurate (gamed) and insecure (tampered) rewards.
In order to focus on tampering problems, we assume accurate feedback and only weaken the secure feedback assumption.

\subsection{REALab vs CFMDPs}
\label{subsec:realab_vs_cfmdp}
REALab and the CFMDP framework are two views on the tampering problem with different strengths.
We can view REALab tasks within the CFMDP framework as described in \S\ref{subsec:interaction_protocol}.
However, we take REALab as our main model because it captures a number of aspects that the CFMDP framework omits in order to remain easy to work with mathematically.

\paragraph{Does the corruption depend on the feedback function?}
When specifying a CFMDP, we choose a fixed corruption function.
It may seem, however, that the corruption function should depend on the agent designer's choice of feedback function.
For example, if the supervisor is provided inaccurate news sources, this may result in high corruption of approval feedback (supervisor approves but would disapprove if they knew better) but low corruption of reward feedback (supervisor's bank balance is unaffected by their inaccurate beliefs).
In principle, both possibilities can be modeled by a single corruption function, since the type of feedback can be indicated in the state.
However, modeling reality with a single corruption function is difficult, because we don't know how the feedback function is implemented in the state: CFMDP states have no internal structure.
Therefore, the environment designer must make careful modeling choices when designing the state space and corruption function.
By contrast, in REALab, there is a natural relationship between the implementation of feedback mechanisms and the state, since registers are physically placed within the state.

\paragraph{Corruption arises from the transition dynamics in REALab}
Related to this, when working with CFMDPs we need to carefully choose assumptions on what kind of corruption is possible, to remain realistic.
In REALab, the environment designer does not have to explicitly choose a corruption function.
Instead, the platform aims to supply a corruption function naturally and implicitly by being grounded in the physics simulation.
This is more similar to how corruption occurs in the real world.

\paragraph{REALab supports corruption of feedback-provider inputs}
In CFMDPs we apply corruption to the output of the feedback function but assume its inputs (the current state and action) arrive uncorrupted.
In REALab, the inputs to the feedback provider are meter register readings and a partial observation.
The meter readings can be corrupted by agent actions, and the observation may misrepresent the underlying state depending on where the agent looks.
These additional opportunities for corruption make REALab more realistic.
They could be modeled in an extension of the CFMDP framework.

\paragraph{REALab has an affordance for increased embeddedness}
In the real world, the abstraction boundary between the feedback provider and the rest of the state is often imperfect.
For example, for a personal assistant receiving feedback via user survey, increasing the user's screen magnification is an action that mostly affects the state and not the feedback mechanism, but if parts of the survey become harder to notice under the new magnification it also affects the feedback.
In REALab, we can embed more of the feedback provider computation, exposing its inner workings to agent actions, by using influenceable components to implement more of that computation.

\subsection{Is REALab a satisfactory unit test for tampering?}
\label{subsec:unit_test}
REALab allows for tampering, while maintaining agent designer flexibility.
In our view, avoiding tampering in REALab is no harder than in the real world.
This judgment is admittedly subjective, though we discuss further below.
To structure this discussion, we consider a variety of approaches for addressing tampering problems,
and whether such approaches might not tamper in REALab but still tamper in the real world, i.e., whether REALab as a unit test has false negatives.

\paragraph{Explicitly detecting and penalizing tampering}
A natural approach to avoiding tampering is to augment the task reward with an auxiliary reward that penalizes tampering.
This is easily implemented in REALab.
One approach is to load a pretrained neural network for detecting tampering into the feedback provider.
In addition to providing an auxiliary reward, the feedback provider might use an additional register to instruct the agent to end the episode when tampering is detected (to prevent further tampering).
A second future possibility would involve adding `tampering detection' meters to REALab (likely with limitations on the number of queries) -- however, this would correspond to assuming that human supervisors are always capable of detecting tampering, with sufficient effort.

The success of this approach depends on detecting all possible forms of tampering.
In general, we believe this is relatively easier in REALab.
Because tampering arises naturally from the simulation physics, REALab allows many different approaches to tampering (e.g., the agent does not need to directly touch a block in order to push it), but we believe this is still a small space of tampering possibilities compared to the real-world examples in \S\ref{subsec:examples}.
This makes REALab a valid but incomplete unit test -- approaches that succeed in REALab may fail in the real world, but if approaches fail in REALab we should be extremely wary of deployment in the real world.

\paragraph{Securing the feedback mechanisms}
In \S\ref{subsec:tampering_likelihood} we discussed how tampering requires somewhat insecure feedback mechanisms.
Therefore, one class of approaches to addressing the tampering problem, sometimes called `boxing the agent', is to isolate the feedback mechanisms from the agent, and add physical or hand-coded barriers such as tripwires to make it difficult for agents to tamper \citep{armstrong2012thinking, Amodei2016concrete,cohen2020asymptotically}.
Boxing is supported in REALab through two mechanisms: specifying locations of REALab components to be more inaccessible, and adding checks against tampering, such as feedback providers checking different registers for tampering.

As before, we believe boxing is likely more difficult in the real world due to the necessity of interacting with human supervisors.
A current limitation of REALab is that registers are less integrated into tasks, relative to human interaction in the real world, making tampering less likely in REALab.
These factors similarly make REALab a valid but incomplete unit test.

\paragraph{Different forms of feedback}
Many proposals to prevent tampering rely on feedback besides instantaneous rewards \citep{everitt2018thesis,HadfieldMenell2016CIRL,Armstrong2017Counterfactual}.
All these algorithms can be empirically studied in REALab, since the semantics of feedback providers and registers are left up to agent designers.
As mentioned in \S\ref{subsec:tampering_likelihood}, we hope to also encourage the use of other forms of feedback and other variations on agent design.
One current limitation on this is bandwidth constraints of the register implementations -- refer to the limitations section below.

\subsection{Current limitations of REALab}
\label{subsec:limitations}

Although REALab has already been used in experiments and to ground discussion of tampering, the platform is still preliminary.
Several important sub-questions remain, such as exactly which registers and meters REALab should support, and how to easily allow agent designers to programatically describe arrangements of REALab components.
To this end, we describe limitations we have encountered in our use of REALab, and suggest possibilities for how they might be addressed by future work.

\paragraph{Register dynamics are disconnected from task dynamics}
An important factor making tampering more likely in many real world applications is that tampering emerges naturally from the intended task -- as discussed under `coupling' in \S\ref{subsec:tampering_likelihood}.
First, the intended task often requires agents to learn predictive models which can also be leveraged for tampering.
In our personal assistant example, this would include the ability to predict the effects of different actions on the user's thoughts and responses.
Second, this makes tampering actions more likely to be explored while learning the task.
In our running example, it is natural for the assistant to make some political content recommendations, some of which will be polarizing.

While REALab aims to emulate this property by reusing the standard simulation dynamics to allow tampering, it is still somewhat looser.
In particular, agents can solve the tasks we have implemented so far without being aware of the registers at all, or the fact that they can be moved.
This created difficulties in several experiments and required us to tune the difficulty of tampering by placing registers closer or further away from reward-providing objects.
When the registers were too far away, the agent would not explore sufficiently to learn about the possibility of tampering.
Future REALab versions could address this limitation by better integrating task and register dynamics -- for example, through studying tasks which themselves rely on moving registers.

\paragraph{Restrictive register implementations}
Our current register implementations are restrictive in two main ways.
First, they have limited bandwidth.
Two-block registers can only communicate a single float, whereas some algorithms, such as Current Reward Function Optimization~\citep{everitt2018thesis}, require communicating large amounts of information between agents and feedback providers.
Second, they only allow fairly restrictive forms of tampering.
For example, two-block registers allow shifting feedback permanently by a constant offset, or changing feedback at the current step, but not richer modifications such as permanently increasing feedback for only specific observations or actions.
These limitations interact: higher bandwidth registers would need to support richer forms of tampering if tampering were to be able to meaningfully affect training.
Both limitations point to a need for new register implementations.

\paragraph{Partially embedded}
Although REALab has an affordance for greater embeddedness (\S\ref{subsec:realab_vs_cfmdp}), it is still a partially embedded platform \citep{Demski2019embedded}.
For example, it does not allow the possiblity of agents manipulating their source code or internal activations.
We believe that forcing communication between agents and feedback providers to be embedded strikes a good balance between computational constraints and modeling the most important tampering problems.
However, future projects might consider embedding more of the computation to study broader forms of tampering.

\section{Related Work}
\label{sec:related_work}

\paragraph{The tampering problem}
The tampering problem is central to several classic problems in the AI safety literature, including the wireheading \citep{ring2011delusion, everitt2016avoiding} and shutdown problems \citep{soares2015corrigibility,hadfield2017off,orseau2016safely,holtman2020agi,holtman2020corrigibility}.
The problem itself has seen some discussion \citep{Everitt2019tampering, Everitt2017CRMDP, leike2017ai, Amodei2016concrete}, where it is sometimes referenced under a broader heading such as reward corruption or reward hacking.
We implicitly adopt the distinction drawn by \citet{Everitt2019tampering} between reward tampering and reward gaming.
While incorrect reward design can also lead to undesirable behavior \citep{Amodei2016concrete,Krakovna2018specification}, we focus solely on tampering.
Similar to us, \citet{everitt2018thesis} and \citet{Demski2019embedded} conceptualize tampering as arising due to learning from embedded, and thus influenceable, feedback.
Most recently \citet{DBLP:conf/ijcai/ArmstrongLOL20} formulate the problem in terms of learning a reward function online.

All the problems above are particularly important when the policy itself is a learned optimizer \citep{duan2016rl,wang2016learning,guez2019investigation,akkaya2019solving}, as noted by \citet{hubinger2019risks}.
Although \citet{hubinger2019risks} focus on so-called inner alignment concerns, similar issues can arise from corrupt feedback.
The corrupt feedback may correlate well with the true feedback on training points, but if the two later diverge during deployment, such as due to distribution shift, a learned optimizer may exhibit dramatically different behavior from what was observed during training.

Many algorithms have been proposed,
including observation-utility maximizers or current reward function optimization \citep{dewey2011learning, everitt2018thesis}, inverse RL-based approaches \citep{HadfieldMenell2016CIRL, hadfield2017inverse}, counterfactual oracles \citep{Armstrong2017Counterfactual}, approval-direction or decoupled approval \citep{Christiano2014approval}, and several approaches for explicitly identifying tampered feedback to avoid training on such data \citep{Everitt2017CRMDP,Mancuso2019Spiky,leike2018scalable}.
Our work is complementary to the papers focused on algorithms: we focus on understanding the tampering problem rather than proposing solutions.
REALab enables evaluating these algorithms for tampering incentives empirically.

\paragraph{Tampering formalisms}
Much of the work focused on algorithms for tampering also introduces specific setups, such as the Off-Switch Game from \citet{hadfield2017inverse} or the Counterfactual Oracle setup from \citep{Armstrong2017Counterfactual}, which can be expressed as particular CFMDPs.
Our CFMDP formalism builds on the Corrupt Reward MDP formalism from \citet{Everitt2017CRMDP}.
As described in \S\ref{subsec:cfmdp}, the primary difference is the use of general feedback functions as opposed to just instantaneous rewards.
More broadly, our formalism highlights the importance of disentangling environment and agent designer roles in order to understand tampering.
This allows modeling a wider range of algorithms -- such as reward-based RL, imitation learning, and value advice RL -- with a common framework and evaluation metric. \\

\noindent \emph{Connection to Bayesian formalisms.}
Bayesian formalisms treat the true reward as a latent variable, and treat observed feedback (often referred to as a proxy reward) as evidence about this latent variable \citep{HadfieldMenell2016CIRL,hadfield2017inverse,woodward2020learning,russell2019human,jeon2020reward}.
While such Bayesian formalisms are typically motivated by the observation that human feedback is imperfect, rather than the possibility of tampering in particular, these problems are closely related -- both concern maximizing an unobserved true reward function based on imperfect observations.

The primary challenge for Bayesian approaches is that the likelihood function $\mathbb{P}[\textrm{observed} \linebreak[1]\textrm{ feedback} \mid \textrm{true reward}]$
must be chosen \textit{a priori} -- it cannot be learned from data because true rewards are latent.
Common choices are that demonstrations are approximately optimal \citep{ziebart2008maximum}, or that proxy rewards produce approximately optimal policies under a proxy MDP \citep{hadfield2017inverse}.
The key assumption is that this likelihood function is accurate, %
otherwise agents may still exploit any difference between the learned posterior and the true reward.

Corruption functions and likelihood functions play similar roles: both relate true feedback to observed feedback.
However, Bayesian formalisms typically commit to a single likelihood, in order to derive a Bayesian inference update, whereas for CFMDPs the agent designer's aim is to find algorithms that succeed for \emph{classes} of corruption functions, e.g. those satisfying various realistic assumptions.
Relatedly, for many common likelihood functions, e.g., \citet{ziebart2008maximum}, the likelihood function does not depend on the current state, whereas state-dependence is a crucial aspect of tampering and corruption functions.
On the other hand, Bayesian formalisms model a broader set of problems since they do not assume accurate feedback as we do.
Furthermore, for many corruption functions, direct application of reward maximization algorithms will not learn an optimal policy, and Bayesian learning algorithms will be more effective. \\

\paragraph{Delineation of environment and agent designer roles}
The distinction we draw between the feedback function and the true reward function is closely related to the preference-parameter confound in \citet{singh2010separating, singh2009rewards}.
Both highlight that while the objective is inherent to the task (preferences), the agent designer can choose the feedback function used for training (parameters).
In particular, our true returns directly correspond to fitness in \citet{singh2009rewards},
and our feedback functions play a similar role to their specific reward functions, but generalize beyond instantaneous rewards.
\citet{jeon2020reward} also highlight the importance of viewing feedback type as a choice of the algorithm designer.
They propose a reward-rational implicit choice framework, which casts a broad array of learning algorithms using different feedback as as Bayesian approaches derived from particular likelihood functions, referred to as \emph{grounding functions} in their work.
We build on both \citet{singh2009rewards} and \citet{jeon2020reward} by considering the presence of corrupt feedback, which we emphasize is a key consideration for designing feedback functions.

\paragraph{Influenceability of human feedback}
An important factor in the prevalence of influenceable feedback is the influenceability of human preferences.
A long empirical literature suggests that stated preferences can be easily influenced, stated preferences can differ significantly from revealed preferences, and that preferences themselves are malleable \citep{tversky1974judgment, tversky1981framing, gilovich2002heuristics}.
This topic is beyond the scope of the current work, but these observations form the basis for behavioral economics \citep{kahneman2003maps,mullainathan2000behavioral}.

Of particular relevance is the degree to which algorithms themselves can influence human preferences.
\citet{Adomavicius2013recommender} show that content recommendation algorithms can significantly influence users' stated enjoyment of films and jokes.
\citet{shi2020effects} study the effectiveness of ML agents for convincing Amazon Mechanical Turk users to make small charity donations.
Probably the largest evidence base focuses on political polarization caused by recommendation algorithms, though even here,
experimental results are limited -- \citet{barbera2015tweeting} partially attributes this to the difficulty of experimentation involving real-world platforms for researchers outside companies operating algorithms on these platforms.
Here, we note that most large-scale experiments to date find the role of recommendation algorithms in contributing to political polarization to be fairly mild, with reported effects in both directions \citep{bakshy2015exposure, flaxman2016filter, barbera2020social}.
Looking more generally, many seemingly mundane decisions in software applications (often partially optimized by algorithms, e.g., in A/B testing) can have large influences on user behavior \citep{anderson2013steering}.

\paragraph{Empirical platforms for tampering}
AI Safety Gridworlds \citep{leike2017ai} study similar problems to REALab.
In their \emph{specification problems}, the agent is trained according to one reward function, but evaluated according to a difference performance function -- these roles are similar to our usage of feedback functions for training and reward functions for evaluation, respectively.
An important difference is that feedback functions in REALab come from the agent designer, not the environment, as discussed in \S\ref{subsec:prior_limitations} -- without this distinction, it is unclear how agents trained to maximize one reward function should learn to maximize a different reward function.

Several other tampering environments are implemented as gridworlds \citep{hadfield2017inverse,majha2019categorizing}, which provide benefits of rapid iteration times.
For REALab, it was important that tampering arise from the underlying physics of the world, as opposed to separate environment dynamics specifically added to model tampering.
While this could also be possible in gridworlds, we found it more straightforward to reuse rigid-body dynamics from an existing physics engine (in our case, Unity).

The platform most similar to REALab is Botworld \citep{soares2014botworld}, which inspired many REALab design choices, such as the focus on registers.
Like REALab, the underlying principle of Botworld is embedded agency, which is taken further -- in Botworld, everything is a register.
Because Botworld agents and environments are both composed of the same limited set of primitives, which share the same dynamics, there is no sharp boundary between agents and their environments.
The primary difference is that REALab is designed for deep RL experimentation and thus operates on higher level primitives.
Similarly, REALab agents are written purely in Python, whereas Botworld agents are written in the constree language in Haskell.

Finally, REALab draws inspiration from a variety of other work demonstrating how environments composed of simple primitives can nonetheless allow complex interactions, beyond just tampering.
Conway's Game of Life is the simplest demonstration in this vein \citep{martin1970mathematical}.
The puzzle game `Baba is You' \citep{babaisyou} is an environment where each level's rules and win conditions are present as physical blocks.
By moving these blocks, players can change how the level works.
In Core War \citep{corewar}, agents are running programs that try to make other agents execute an illegal instruction -- because the source code and working memory for all agents are located in a shared memory space, agents can modify the execution of other programs as well as their own.

\section{Conclusion}
\label{sec:conclusion}
Many important problems can be posed as tampering problems (\S\ref{subsec:personal_assistant}, \S\ref{subsec:examples}).
However, tampering problems lie outside the scope of commonly used RL environments and the MDP formalism, due to their assumption of uninfluenceable feedback functions (\S\ref{subsec:prior_limitations}).
We hope readers leave this paper with two main takeaways.
First, the tampering problem is well-posed -- it can be studied formally (\S\ref{subsec:cfmdp}) and empirically (\S\ref{subsec:realab}), allowing the agent designer to make algorithmic choices to account for the possibility of tampering.
Second, REALab provides an empirical platform for studying tampering problems in simulation.
In particular, we aim for REALab to provide a unit test for tampering (\S\ref{subsec:unit_test}) -- if we can't make algorithms work in REALab, we should be very hesitant before applying them to real world applications with the potential for tampering.

\section*{Acknowledgements}
We would like to especially thank Paul Christiano, Evan Hubinger, and Rohin Shah for extensive discussions in developing and refining these ideas.
We are also grateful to
Frederic Besse, Andrew Bolt, Orlagh Burns, Charlie Deck, Dario de Cesare, Koen Holtman, Geoffrey Irving, Zac Kenton, Pushmeet Kohli, Jan Leike, Nat McAleese, Vladimir Mikulik, Matthew Rahtz, Jason Sanmiya, Adam Shimi, Guy Simmons, and Alex Zhu
for many helpful discussions throughout the course of this work.

\bibliographystyle{plainnat}
\bibliography{refs,realab_refs}

\begin{thebibliography}{67}
\providecommand{\natexlab}[1]{#1}
\providecommand{\url}[1]{\texttt{#1}}
\expandafter\ifx\csname urlstyle\endcsname\relax
  \providecommand{\doi}[1]{doi: #1}\else
  \providecommand{\doi}{doi: \begingroup \urlstyle{rm}\Url}\fi

\bibitem[Adomavicius et~al.(2013)Adomavicius, Bockstedt, Curley, and
  Zhang]{Adomavicius2013recommender}
Gediminas Adomavicius, Jesse~C. Bockstedt, Shawn~P. Curley, and Jingjing Zhang.
\newblock Do recommender systems manipulate consumer preferences? {A} study of
  anchoring effects.
\newblock \emph{Information Systems Research}, 24\penalty0 (4):\penalty0
  956--975, 2013.

\bibitem[Amodei et~al.(2016)Amodei, Olah, Steinhardt, Christiano, Schulman, and
  Man{\'{e}}]{Amodei2016concrete}
Dario Amodei, Chris Olah, Jacob Steinhardt, Paul~F. Christiano, John Schulman,
  and Dan Man{\'{e}}.
\newblock Concrete problems in {AI} safety.
\newblock \emph{CoRR}, 1606.06565, 2016.

\bibitem[Anderson et~al.(2013)Anderson, Huttenlocher, Kleinberg, and
  Leskovec]{anderson2013steering}
Ashton Anderson, Daniel Huttenlocher, Jon Kleinberg, and Jure Leskovec.
\newblock Steering user behavior with badges.
\newblock In \emph{Proceedings of the 22nd international conference on World
  Wide Web}, pages 95--106, 2013.

\bibitem[Armstrong and O'Rourke(2017)]{Armstrong2017Counterfactual}
Stuart Armstrong and Xavier O'Rourke.
\newblock Good and safe uses of {AI} oracles.
\newblock \emph{CoRR}, abs/1711.05541, 2017.

\bibitem[Armstrong et~al.(2012)Armstrong, Sandberg, and
  Bostrom]{armstrong2012thinking}
Stuart Armstrong, Anders Sandberg, and Nick Bostrom.
\newblock Thinking inside the box: Controlling and using an oracle ai.
\newblock \emph{Minds and Machines}, 22\penalty0 (4):\penalty0 299--324, 2012.

\bibitem[Armstrong et~al.(2020)Armstrong, Leike, Orseau, and
  Legg]{DBLP:conf/ijcai/ArmstrongLOL20}
Stuart Armstrong, Jan Leike, Laurent Orseau, and Shane Legg.
\newblock Pitfalls of learning a reward function online.
\newblock In Christian Bessiere, editor, \emph{Proceedings of the Twenty-Ninth
  International Joint Conference on Artificial Intelligence, {IJCAI} 2020
  [scheduled for July 2020, Yokohama, Japan, postponed due to the Corona
  pandemic]}, pages 1592--1600. ijcai.org, 2020.
\newblock \doi{10.24963/ijcai.2020/221}.
\newblock URL \url{https://doi.org/10.24963/ijcai.2020/221}.

\bibitem[Bakshy et~al.(2015)Bakshy, Messing, and Adamic]{bakshy2015exposure}
Eytan Bakshy, Solomon Messing, and Lada~A Adamic.
\newblock Exposure to ideologically diverse news and opinion on facebook.
\newblock \emph{Science}, 348\penalty0 (6239):\penalty0 1130--1132, 2015.

\bibitem[Barber{\'a}(2020)]{barbera2020social}
Pablo Barber{\'a}.
\newblock Social media, echo chambers, and political polarization.
\newblock \emph{Social Media and Democracy: The State of the Field, Prospects
  for Reform}, page~34, 2020.

\bibitem[Barber{\'a} et~al.(2015)Barber{\'a}, Jost, Nagler, Tucker, and
  Bonneau]{barbera2015tweeting}
Pablo Barber{\'a}, John~T Jost, Jonathan Nagler, Joshua~A Tucker, and Richard
  Bonneau.
\newblock Tweeting from left to right: Is online political communication more
  than an echo chamber?
\newblock \emph{Psychological science}, 26\penalty0 (10):\penalty0 1531--1542,
  2015.

\bibitem[Barto and Mahadevan(2003)]{barto2003recent}
Andrew~G Barto and Sridhar Mahadevan.
\newblock Recent advances in hierarchical reinforcement learning.
\newblock \emph{Discrete event dynamic systems}, 13\penalty0 (1-2):\penalty0
  41--77, 2003.

\bibitem[Beattie et~al.(2016)Beattie, Leibo, Teplyashin, Ward, Wainwright,
  K{\"u}ttler, Lefrancq, Green, Vald{\'e}s, Sadik, et~al.]{beattie2016deepmind}
Charles Beattie, Joel~Z Leibo, Denis Teplyashin, Tom Ward, Marcus Wainwright,
  Heinrich K{\"u}ttler, Andrew Lefrancq, Simon Green, V{\'\i}ctor Vald{\'e}s,
  Amir Sadik, et~al.
\newblock {DeepMind Lab}.
\newblock \emph{arXiv preprint arXiv:1612.03801}, 2016.

\bibitem[Bostrom(2014)]{Bostrom2016Superintelligence}
Nick Bostrom.
\newblock \emph{Superintelligence: Paths, Dangers, Strategies}.
\newblock Oxford University Press, Inc., USA, 2014.
\newblock ISBN 0198739834.

\bibitem[Brockman et~al.(2016)Brockman, Cheung, Pettersson, Schneider,
  Schulman, Tang, and Zaremba]{brockman2016openai}
Greg Brockman, Vicki Cheung, Ludwig Pettersson, Jonas Schneider, John Schulman,
  Jie Tang, and Wojciech Zaremba.
\newblock Openai gym.
\newblock \emph{arXiv preprint arXiv:1606.01540}, 2016.

\bibitem[Christiano(2014)]{Christiano2014approval}
Paul Christiano.
\newblock Approval-directed agents.
\newblock AI Alignment on Medium, 2014.

\bibitem[Cohen et~al.(2020)Cohen, Vellambi, and
  Hutter]{cohen2020asymptotically}
Michael~K Cohen, Badri~N Vellambi, and Marcus Hutter.
\newblock Asymptotically unambitious artificial general intelligence.
\newblock In \emph{AAAI}, pages 2467--2476, 2020.

\bibitem[Daswani et~al.(2014)Daswani, Sunehag, and
  Hutter]{Daswani2014ValueAdvice}
Mayank Daswani, Peter Sunehag, and Marcus Hutter.
\newblock Reinforcement learning with value advice.
\newblock In \emph{{ACML}}, volume~39 of \emph{{JMLR} Workshop and Conference
  Proceedings}. JMLR.org, 2014.

\bibitem[Demski and Garrabrant(2019)]{Demski2019embedded}
Abram Demski and Scott Garrabrant.
\newblock Embedded agency.
\newblock \emph{CoRR}, 1902.09469, 2019.

\bibitem[Dewdney(1984)]{corewar}
Alexander~Keewatin Dewdney.
\newblock Core war, 1984.
\newblock URL \url{corewar.co.uk}.

\bibitem[Dewey(2011)]{dewey2011learning}
Daniel Dewey.
\newblock Learning what to value.
\newblock In \emph{International Conference on Artificial General
  Intelligence}, pages 309--314. Springer, 2011.

\bibitem[Duan et~al.(2016)Duan, Schulman, Chen, Bartlett, Sutskever, and
  Abbeel]{duan2016rl}
Yan Duan, John Schulman, Xi~Chen, Peter~L Bartlett, Ilya Sutskever, and Pieter
  Abbeel.
\newblock Rl2: Fast reinforcement learning via slow reinforcement learning.
\newblock \emph{arXiv preprint arXiv:1611.02779}, 2016.

\bibitem[Everitt(2018)]{everitt2018thesis}
Tom Everitt.
\newblock \emph{Towards Safe Artificial General Intelligence}.
\newblock PhD thesis, Australian National University, May 2018.
\newblock URL \url{http://hdl.handle.net/1885/164227}.

\bibitem[Everitt and Hutter(2016)]{everitt2016avoiding}
Tom Everitt and Marcus Hutter.
\newblock Avoiding wireheading with value reinforcement learning.
\newblock In \emph{International Conference on Artificial General
  Intelligence}, pages 12--22. Springer, 2016.

\bibitem[Everitt and Hutter(2019)]{Everitt2019tampering}
Tom Everitt and Marcus Hutter.
\newblock Reward tampering problems and solutions in reinforcement learning.
\newblock \emph{CoRR}, 1908.04734, 2019.

\bibitem[Everitt et~al.(2017)Everitt, Krakovna, Orseau, Hutter, and
  Legg]{Everitt2017CRMDP}
Tom Everitt, Victoria Krakovna, Laurent Orseau, Marcus Hutter, and Shane Legg.
\newblock Reinforcement learning with a corrupted reward channel.
\newblock \emph{International Joint Conferences on Artificial Intelligence},
  2017.

\bibitem[Flaxman et~al.(2016)Flaxman, Goel, and Rao]{flaxman2016filter}
Seth Flaxman, Sharad Goel, and Justin~M Rao.
\newblock Filter bubbles, echo chambers, and online news consumption.
\newblock \emph{Public opinion quarterly}, 80\penalty0 (S1):\penalty0 298--320,
  2016.

\bibitem[Gilovich et~al.(2002)Gilovich, Griffin, and
  Kahneman]{gilovich2002heuristics}
Thomas Gilovich, Dale Griffin, and Daniel Kahneman.
\newblock \emph{Heuristics and biases: The psychology of intuitive judgment}.
\newblock Cambridge university press, 2002.

\bibitem[Guez et~al.(2019)Guez, Mirza, Gregor, Kabra, Racaniere, Weber, Raposo,
  Santoro, Orseau, Eccles, et~al.]{guez2019investigation}
Arthur Guez, Mehdi Mirza, Karol Gregor, Rishabh Kabra, S{\'e}bastien Racaniere,
  Th{\'e}ophane Weber, David Raposo, Adam Santoro, Laurent Orseau, Tom Eccles,
  et~al.
\newblock An investigation of model-free planning.
\newblock \emph{arXiv preprint arXiv:1901.03559}, 2019.

\bibitem[Gulcehre et~al.(2020)Gulcehre, Wang, Novikov, Paine, Colmenarejo,
  Zolna, Agarwal, Merel, Mankowitz, Paduraru, et~al.]{gulcehre2020rl}
Caglar Gulcehre, Ziyu Wang, Alexander Novikov, Tom~Le Paine, Sergio~G{\'o}mez
  Colmenarejo, Konrad Zolna, Rishabh Agarwal, Josh Merel, Daniel Mankowitz,
  Cosmin Paduraru, et~al.
\newblock Rl unplugged: Benchmarks for offline reinforcement learning.
\newblock \emph{arXiv preprint arXiv:2006.13888}, 2020.

\bibitem[Hadfield-Menell et~al.(2016)Hadfield-Menell, Russell, Abbeel, and
  Dragan]{HadfieldMenell2016CIRL}
Dylan Hadfield-Menell, Stuart~J Russell, Pieter Abbeel, and Anca Dragan.
\newblock Cooperative inverse reinforcement learning.
\newblock In \emph{Advances in neural information processing systems}, pages
  3909--3917, 2016.

\bibitem[Hadfield-Menell et~al.(2017{\natexlab{a}})Hadfield-Menell, Dragan,
  Abbeel, and Russell]{hadfield2017off}
Dylan Hadfield-Menell, Anca Dragan, Pieter Abbeel, and Stuart Russell.
\newblock The off-switch game.
\newblock In \emph{Workshops at the Thirty-First AAAI Conference on Artificial
  Intelligence}, 2017{\natexlab{a}}.

\bibitem[Hadfield-Menell et~al.(2017{\natexlab{b}})Hadfield-Menell, Milli,
  Abbeel, Russell, and Dragan]{hadfield2017inverse}
Dylan Hadfield-Menell, Smitha Milli, Pieter Abbeel, Stuart~J Russell, and Anca
  Dragan.
\newblock Inverse reward design.
\newblock In \emph{Advances in neural information processing systems}, pages
  6765--6774, 2017{\natexlab{b}}.

\bibitem[Holtman(2020{\natexlab{a}})]{holtman2020agi}
Koen Holtman.
\newblock Agi agent safety by iteratively improving the utility function.
\newblock \emph{arXiv preprint arXiv:2007.05411}, 2020{\natexlab{a}}.

\bibitem[Holtman(2020{\natexlab{b}})]{holtman2020corrigibility}
Koen Holtman.
\newblock Corrigibility with utility preservation.
\newblock \emph{arXiv preprint arXiv:1908.01695}, 2020{\natexlab{b}}.

\bibitem[Hubinger et~al.(2019)Hubinger, van Merwijk, Mikulik, Skalse, and
  Garrabrant]{hubinger2019risks}
Evan Hubinger, Chris van Merwijk, Vladimir Mikulik, Joar Skalse, and Scott
  Garrabrant.
\newblock Risks from learned optimization in advanced machine learning systems.
\newblock \emph{arXiv preprint arXiv:1906.01820}, 2019.

\bibitem[Jeon et~al.(2020)Jeon, Milli, and Dragan]{jeon2020reward}
Hong~Jun Jeon, Smitha Milli, and Anca~D Dragan.
\newblock Reward-rational (implicit) choice: A unifying formalism for reward
  learning.
\newblock \emph{arXiv preprint arXiv:2002.04833}, 2020.

\bibitem[Kahneman(2003)]{kahneman2003maps}
Daniel Kahneman.
\newblock Maps of bounded rationality: Psychology for behavioral economics.
\newblock \emph{American economic review}, 93\penalty0 (5):\penalty0
  1449--1475, 2003.

\bibitem[Knox and Stone(2008)]{knox2008tamer}
W~Bradley Knox and Peter Stone.
\newblock {TAMER}: Training an agent manually via evaluative reinforcement.
\newblock In \emph{2008 7th IEEE International Conference on Development and
  Learning}, pages 292--297. IEEE, 2008.

\bibitem[Krakovna et~al.(2020)Krakovna, Uesato, Mikulik, Rahtz, Everitt, Kumar,
  Kenton, Leike, and Legg]{Krakovna2018specification}
Victoria Krakovna, Jonathan Uesato, Vladimir Mikulik, Matthew Rahtz, Tom
  Everitt, Ramana Kumar, Zac Kenton, Jan Leike, and Shane Legg.
\newblock {Specification gaming: the flip side of AI ingenuity}.
\newblock DeepMind Blog, 2020.

\bibitem[Leike et~al.(2017)Leike, Martic, Krakovna, Ortega, Everitt, Lefrancq,
  Orseau, and Legg]{leike2017ai}
Jan Leike, Miljan Martic, Victoria Krakovna, Pedro~A Ortega, Tom Everitt,
  Andrew Lefrancq, Laurent Orseau, and Shane Legg.
\newblock Ai safety gridworlds.
\newblock \emph{arXiv preprint arXiv:1711.09883}, 2017.

\bibitem[Leike et~al.(2018)Leike, Krueger, Everitt, Martic, Maini, and
  Legg]{leike2018scalable}
Jan Leike, David Krueger, Tom Everitt, Miljan Martic, Vishal Maini, and Shane
  Legg.
\newblock Scalable agent alignment via reward modeling: a research direction.
\newblock \emph{arXiv preprint arXiv:1811.07871}, 2018.

\bibitem[Levine et~al.(2020)Levine, Kumar, Tucker, and Fu]{levine2020offline}
Sergey Levine, Aviral Kumar, George Tucker, and Justin Fu.
\newblock Offline reinforcement learning: Tutorial, review, and perspectives on
  open problems.
\newblock \emph{arXiv preprint arXiv:2005.01643}, 2020.

\bibitem[Majha et~al.(2019)Majha, Sarkar, and Zagami]{majha2019categorizing}
Arushi Majha, Sayan Sarkar, and Davide Zagami.
\newblock Categorizing wireheading in partially embedded agents.
\newblock \emph{arXiv preprint arXiv:1906.09136}, 2019.

\bibitem[Mancuso et~al.(2019)Mancuso, Kisielewski, Lindner, and
  Singh]{Mancuso2019Spiky}
Jason Mancuso, Tomasz Kisielewski, David Lindner, and Alok Singh.
\newblock Detecting spiky corruption in {Markov} decision processes.
\newblock In \emph{AISafety@IJCAI}, volume 2419 of \emph{{CEUR} Workshop
  Proceedings}. CEUR-WS.org, 2019.

\bibitem[Martin(1970)]{martin1970mathematical}
Gardner Martin.
\newblock Mathematical games: The fantastic combinations of john conway’s new
  solitaire game” life”.
\newblock \emph{Scientific American}, 223:\penalty0 120--123, 1970.

\bibitem[Merel et~al.(2018)Merel, Hasenclever, Galashov, Ahuja, Pham, Wayne,
  Teh, and Heess]{merel2018neural}
Josh Merel, Leonard Hasenclever, Alexandre Galashov, Arun Ahuja, Vu~Pham, Greg
  Wayne, Yee~Whye Teh, and Nicolas Heess.
\newblock Neural probabilistic motor primitives for humanoid control.
\newblock \emph{arXiv preprint arXiv:1811.11711}, 2018.

\bibitem[Mullainathan and Thaler(2000)]{mullainathan2000behavioral}
Sendhil Mullainathan and Richard~H Thaler.
\newblock Behavioral economics.
\newblock Technical report, National Bureau of Economic Research, 2000.

\bibitem[OpenAI et~al.(2019)OpenAI, Akkaya, Andrychowicz, Chociej, Litwin,
  McGrew, Petron, Paino, Plappert, Powell, Ribas, et~al.]{akkaya2019solving}
OpenAI, Ilge Akkaya, Marcin Andrychowicz, Maciek Chociej, Mateusz Litwin, Bob
  McGrew, Arthur Petron, Alex Paino, Matthias Plappert, Glenn Powell, Raphael
  Ribas, et~al.
\newblock Solving rubik's cube with a robot hand.
\newblock \emph{arXiv preprint arXiv:1910.07113}, 2019.

\bibitem[Orseau and Armstrong(2016)]{orseau2016safely}
Laurent Orseau and Stuart Armstrong.
\newblock Safely interruptible agents.
\newblock \emph{Association for Uncertainty in Artificial Intelligence}, 2016.

\bibitem[Reed and De~Freitas(2015)]{reed2015neural}
Scott Reed and Nando De~Freitas.
\newblock Neural programmer-interpreters.
\newblock \emph{arXiv preprint arXiv:1511.06279}, 2015.

\bibitem[Ring and Orseau(2011)]{ring2011delusion}
Mark Ring and Laurent Orseau.
\newblock Delusion, survival, and intelligent agents.
\newblock In \emph{International Conference on Artificial General
  Intelligence}, pages 11--20. Springer, 2011.

\bibitem[Ross et~al.(2011)Ross, Gordon, and Bagnell]{ross2011reduction}
St{\'e}phane Ross, Geoffrey Gordon, and Drew Bagnell.
\newblock A reduction of imitation learning and structured prediction to
  no-regret online learning.
\newblock In \emph{Proceedings of the fourteenth international conference on
  artificial intelligence and statistics}, pages 627--635, 2011.

\bibitem[Russell(2019)]{russell2019human}
Stuart Russell.
\newblock \emph{Human compatible: Artificial intelligence and the problem of
  control}.
\newblock Penguin, 2019.

\bibitem[Shi et~al.(2020)Shi, Wang, Oh, Zhang, Sahay, and Yu]{shi2020effects}
Weiyan Shi, Xuewei Wang, Yoo~Jung Oh, Jingwen Zhang, Saurav Sahay, and Zhou Yu.
\newblock Effects of persuasive dialogues: Testing bot identities and inquiry
  strategies.
\newblock In \emph{Proceedings of the 2020 CHI Conference on Human Factors in
  Computing Systems}, pages 1--13, 2020.

\bibitem[Singh et~al.(2009)Singh, Lewis, and Barto]{singh2009rewards}
Satinder Singh, Richard~L Lewis, and Andrew~G Barto.
\newblock Where do rewards come from.
\newblock In \emph{Proceedings of the annual conference of the cognitive
  science society}, pages 2601--2606. Cognitive Science Society, 2009.

\bibitem[Singh et~al.(2010)Singh, Lewis, Sorg, Barto, and
  Helou]{singh2010separating}
Satinder Singh, Richard~L Lewis, Jonathan Sorg, A~Barto, and Akram Helou.
\newblock On separating agent designer goals from agent goals: Breaking the
  preferences--parameters confound, 2010.

\bibitem[Soares and Fallenstein(2014)]{soares2014botworld}
Nate Soares and Benja Fallenstein.
\newblock Botworld 1.0 (technical report).
\newblock 2014.

\bibitem[Soares et~al.(2015)Soares, Fallenstein, Armstrong, and
  Yudkowsky]{soares2015corrigibility}
Nate Soares, Benja Fallenstein, Stuart Armstrong, and Eliezer Yudkowsky.
\newblock Corrigibility.
\newblock In \emph{Workshops at the Twenty-Ninth AAAI Conference on Artificial
  Intelligence}, 2015.

\bibitem[Sutton and Barto(1998)]{sutton1998introduction}
Richard~S Sutton and Andrew~G Barto.
\newblock \emph{Reinforcement learning: an introduction}.
\newblock MIT Press Cambridge, 1998.

\bibitem[Sutton et~al.(1999)Sutton, Precup, and Singh]{sutton1999between}
Richard~S Sutton, Doina Precup, and Satinder Singh.
\newblock Between mdps and semi-mdps: A framework for temporal abstraction in
  reinforcement learning.
\newblock \emph{Artificial intelligence}, 112\penalty0 (1-2):\penalty0
  181--211, 1999.

\bibitem[Teikari(2019)]{babaisyou}
Arvi Teikari.
\newblock Baba is you, 2019.
\newblock URL \url{https://hempuli.com/baba/}.

\bibitem[Tversky and Kahneman(1974)]{tversky1974judgment}
Amos Tversky and Daniel Kahneman.
\newblock Judgment under uncertainty: Heuristics and biases.
\newblock \emph{science}, 185\penalty0 (4157):\penalty0 1124--1131, 1974.

\bibitem[Tversky and Kahneman(1981)]{tversky1981framing}
Amos Tversky and Daniel Kahneman.
\newblock The framing of decisions and the psychology of choice.
\newblock \emph{science}, 211\penalty0 (4481):\penalty0 453--458, 1981.

\bibitem[Uesato et~al.(2020)Uesato, Kumar, Krakovna, Everitt, Ngo, and
  Legg]{kumar2020da}
Jonathan Uesato, Ramana Kumar, Victoria Krakovna, Tom Everitt, Richard Ngo, and
  Shane Legg.
\newblock Aligned tampering incentives for deep reinforcement learning.
\newblock \emph{arXiv}, 2020.

\bibitem[Unity(2018)]{unity}
Unity.
\newblock Unity, 2018.
\newblock URL \url{http://unity3d.com}.

\bibitem[Wang et~al.(2016)Wang, Kurth-Nelson, Tirumala, Soyer, Leibo, Munos,
  Blundell, Kumaran, and Botvinick]{wang2016learning}
Jane~X Wang, Zeb Kurth-Nelson, Dhruva Tirumala, Hubert Soyer, Joel~Z Leibo,
  Remi Munos, Charles Blundell, Dharshan Kumaran, and Matt Botvinick.
\newblock Learning to reinforcement learn.
\newblock \emph{arXiv preprint arXiv:1611.05763}, 2016.

\bibitem[Woodward et~al.(2020)Woodward, Finn, and
  Hausman]{woodward2020learning}
Mark Woodward, Chelsea Finn, and Karol Hausman.
\newblock Learning to interactively learn and assist.
\newblock In \emph{AAAI}, pages 2535--2543, 2020.

\bibitem[Ziebart et~al.(2008)Ziebart, Maas, Bagnell, and
  Dey]{ziebart2008maximum}
Brian~D Ziebart, Andrew~L Maas, J~Andrew Bagnell, and Anind~K Dey.
\newblock Maximum entropy inverse reinforcement learning.
\newblock In \emph{Aaai}, volume~8, pages 1433--1438. Chicago, IL, USA, 2008.

\end{thebibliography}
\appendix
\section{Glossary}

\begin{description}
  \item[Accurate feedback]
    Feedback such that behaving according to the feedback corresponds to doing the intended task.
    For example, reward feedback is accurate if any behavior that maximizes the (true) return is desired.
    This is to contrast with inaccurate proxy feedback, for which there may be degenerate behaviors that are good according to the feedback but actually undesirable.
  \item[Agent]
    An agent is an instance of an agent design, e.g., a sequence of policies produced by the learning algorithm using data from interaction with the environment and feedback function.
  \item[Agent design]
    A choice of a learning algorithm and a feedback function, which together specify an agent designed to solve a particular task or set of tasks.
    For a REALab agent, the design also includes the arrangement of any REALab components.
  \item[Agent designer]
    The role in RL that is responsible for specifying an agent design (learning algorithm and feedback function), to contrast with the environment designer who specifies the task and evaluation metric.
  \item[Corruption]
    A change to the value of some variable, to the output of some function, or to the machinery or mental state that implements some function, that results in observed values being incorrect.
    See Corrupt feedback for a more formal definition of that instance.
  \item[Corrupt feedback]
    Observed feedback that differs from the corresponding true feedback.
  \item[Corruption function]
    In CFMDPs, the corruption function is $c$, taking the true feedback, current query, and next state as input, and producing the corrupted feedback.
    In REALab, the corruption function arises from the environment dynamics, and represents the difference in the agent's observed feedback compared to the true feedback from the feedback provider.
  \item[Embedded]
    Part of the same causally connected world. Influenceable.
  \item[Embedded agency]
    The study of agents that are embedded in their environments.
    Picking apart the abstraction boundary between agent and environment in standard RL frameworks can illuminate problems relevant to real-world agent deployment such as the tampering problem.
  \item[Embedded feedback]
    An RL environment has embedded feedback if the mechanisms for providing feedback to the agent are influenceable.
    In other words, the agent's observed feedback depends not just on the feedback provided but also on the environment state.
  \item[Environment designer]
    The role in RL that is responsible for specifying a task including the evaluation metric (true reward function), but not including the feedback function (which is the responsibility of the agent designer).
  \item[Evaluation metric]
    The metric that defines performance of different agents/policies on a task.
    This metric may be infeasible to compute as a feedback signal for training agents.
  \item[Feedback]
    Information provided to an agent to help it learn the intended task.
    Feedback generalizes the notion of reward in standard MDPs.
    Examples include rewards, demonstrations, and value advice.
  \item[Feedback function]
    In CFMDPs, the feedback function is $\delta$, taking a state and query as input and producing feedback as output.
    Analogously in REALab, the feedback function is the function implemented by a feedback provider, taking an observation and register readings as input and producing feedback as output.
  \item[Feedback provider]
    A REALab component that implements a feedback function.
    Feedback providers receive the agent observation and any number of register readings as input, including from meter registers, and write to any number of registers.
  \item[Gaming]
    Behaviors that exploit inaccurate feedback, doing well according to the feedback function but not succeeding at the intended task.
    Gaming focuses on deviations between the feedback function and the intended task, whereas tampering focuses on deviations between true and observed feedback
  \item[Influenceable]
    Can be corrupted, especially as a result of actions taken by an agent.
  \item[Learning algorithm]
    An algorithm for updating a policy given a history of interaction data (observations, actions, queries, and observed feedback).
  \item[Meter]
    A REALab component that measures some feature of the environment state.
    Meter registers can be used directly by feedback providers, but not agents.
    Any function of the underlying simulator state whose output can be represented by a register can be realized as a meter.
  \item[Observed feedback]
    The feedback information that arrives as input to an agent's learning algorithm.
    This is a potentially corrupted version of the output of the feedback provider.
  \item[Policy]
    A (stochastic) map from observations to actions and/or queries.
  \item[Query]
    An input (in addition to the current observation) to the feedback function, which is supplied by the agent.
  \item[Register]
    A (simulated, embedded) physical object used to store information.
  \item[Reward]
    The usual feedback information in RL: a real number provided to the agent at each time step.
    The usual way to specify a task in RL is as maximization of the discounted sum of rewards (called the return).
  \item[Secure feedback]
    Feedback that is uninfluenceable, for example, by not being embedded.
    Feedback produced by an embedded process may be more or less secure depending on how easily it can be corrupted.
  \item[True feedback]
    The output of the feedback function, without any corruption.
  \item[True reward]
    Uncorrupted output of the true reward function, which is defined by the environment designer.
    True returns are the metric for evaluating the performance of agents.
  \item[Tampering]
    Informally, agent behavior that causes corruption of its feedback or other observations. In this work, we do not distinguish between tampering and accidental corruption for individual cases, but in aggregate, we say that agents tamper if they produce higher corruption levels than baseline agents.
  \item[Task]
    A REALab task corresponds to a single CFMDP instance.
    This is analogous to a level in DMLab \citep{beattie2016deepmind} or environment in OpenAI Gym \citep{brockman2016openai}.
  \item[Uninfluenceable]
    See influenceable.
\end{description}

\end{document}